\documentclass{article}
\usepackage{acra}

\usepackage{graphics} 
\usepackage{graphicx} 
\usepackage{subfigure}
\usepackage{epsfig} 
\usepackage{amsmath} 
\usepackage{amssymb}  
\usepackage[linesnumbered, lined, ruled]{algorithm2e}
\usepackage{epstopdf}
\usepackage{bm}
\usepackage{hyperref}
\usepackage{url}
\usepackage{stfloats}
\usepackage{array}
\usepackage[T1]{fontenc}
\usepackage{booktabs}
\usepackage[table]{xcolor}
\usepackage{lastpage}
\usepackage{multirow}
\usepackage{todonotes}

\begin{document}

\title{Towards Vision-Based Deep Reinforcement Learning \\ for Robotic Motion Control}

\author{
  Fangyi Zhang, J\"urgen Leitner, Michael Milford, Ben Upcroft, Peter Corke\\
	ARC Centre of Excellence for Robotic Vision (ACRV)\\ 
      Queensland University of Technology (QUT)\\
    fangyi.zhang@hdr.qut.edu.au
}

\maketitle  
\begin{abstract}

This paper introduces a machine learning based system for controlling a robotic manipulator with visual perception only. The capability to autonomously learn robot controllers solely from raw-pixel images and without any prior knowledge of configuration is shown for the first time. We build upon the success of recent deep reinforcement learning and develop a system for learning target reaching with a three-joint robot manipulator using external visual observation. A Deep Q Network (DQN) was demonstrated to perform target reaching after training in simulation. Transferring the network to real hardware and real observation in a naive approach failed, but experiments show that the network works when replacing camera images with synthetic images.

\end{abstract}

\section{Introduction}
\label{sec:intro}
Robots are widely used to complete various manipulation tasks in industrial manufacturing factories where environments are relatively static and simple. 
However, these operations are still challenging for robots in highly dynamic and complex environments commonly encountered in everyday life. 
Nevertheless, humans are able to manipulate in such highly dynamic and complex environments.
We seem to be able to learn manipulation skills by observing how others perform them (learning from observation), as well as, master new skills through trial and error (learning from exploration). Inspired by this, we want robots to learn and master manipulation skills in the same way. 

\begin{figure}[t]
  \centering
  \includegraphics[width=0.95\columnwidth]{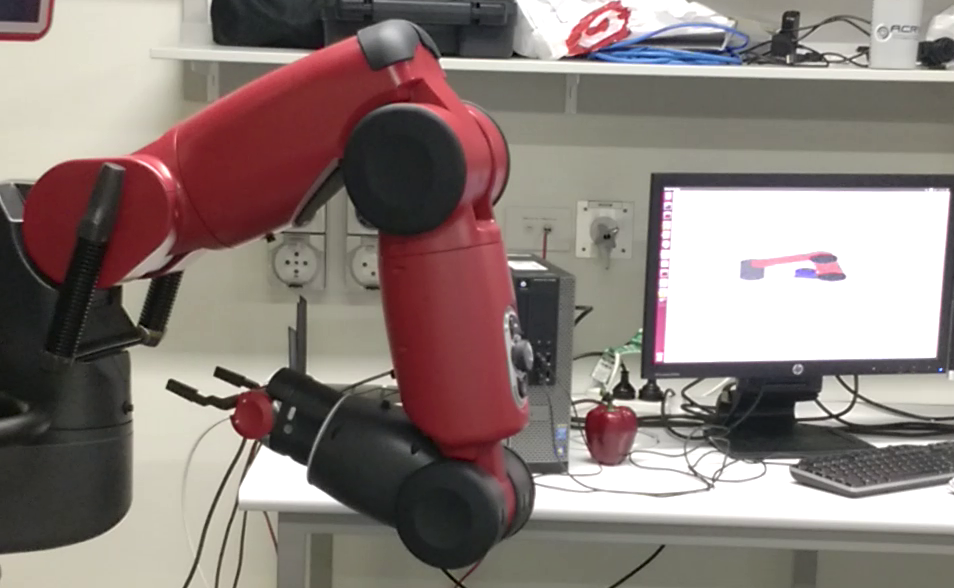}
\caption{Baxter's arm being controlled by a trained deep Q Network (DQN). Synthetic images (on the right) are fed into the DQN to overcome some of the real-world issues encountered, i.e., the differences between training and testing settings.} \label{fig:baxter}
\end{figure}

To give robots the ability to learn from exploration, methods are required that are able to learn autonomously and which are flexible to a range of differing manipulation tasks. A promising candidate for autonomous learning in this regard is Deep Reinforcement Learning (DRL), which combines reinforcement learning and deep learning. One topical example of DRL is the Deep Q Network (DQN), which, after learning to play Atari 2600 games over 38 days, was able to match human performance when playing the game \cite{mnih2013playing,mnih2015human}. Despite their promise, applying DQNs to "perfect" and relatively simple computer game worlds is a far cry from deploying them in complex robotic manipulation tasks, especially when factors such as sensor noise and image offsets are considered.

This paper takes the first steps towards enabling DQNs to be used for learning robotic manipulation. We focus on learning these skills from visual observation of the manipulator, without any prior knowledge of configuration or joint state. Towards this end, as first steps, we assess the feasibility of using DQNs to perform a simple target reaching task, an important component of general manipulation tasks such as object picking. In particular, we make the following contributions:

\begin{itemize}
\item We present a DQN-based learning system for a target reaching task. The system consists of three components: a 2D robotic arm simulator for target reaching, a DQN learner, and ROS-based interfaces to enable operation on a Baxter robot.

\item We train agents in simulation and evaluate them in both simulation and real-world target reaching experiments. The experiments in simulation are conducted with varying levels of noise, image offsets, initial arm poses and link lengths, which are common concerns in robotic motion control and manipulation.

\item We identify and discuss a number of issues and opportunities for future work towards enabling vision-based deep reinforcement learning in real-world robotic manipulation.

\end{itemize}

\section{Related Work}
\label{sec:related_work}

\subsection{Vision-based Robotic Manipulation}
Vision-based robotic manipulation is the process by which robots use their manipulators (such as robotic arms) to rearrange environments \cite{mason2001mechanics}, based on camera images.
The early vision-based robotic manipulation was implemented using pose-based (position and orientation) closed-loop control, where vision was typically used to extract the pose of an object as an input for a manipulation controller at the beginning of a task \cite{kragic2002survey}.

Most current vision-based robotic manipulation methods are closed-loop based on visual perception. 
A vision-based manipulation system was implemented on a Johns Hopkins ``Steady Hand Robot'' for cooperative manipulation at millimeter to micrometer scales, using virtual fixtures \cite{bettini2004vision}. With both monocular and binocular vision cues, various closed-loop visual strategies were applied to enable robots to manipulate both known and unknown objects \cite{kragic2005vision}. 

Also, various learning methods have been applied to implement complex manipulation tasks in the real world. With continuous hidden Markov models (HMMs), a humanoid robot was able to learn dual-arm manipulation tasks from human demonstrations through vision \cite{asfour2008imitation}. However, most of these algorithms are for specific tasks and need much prior knowledge. They are not flexible for learning a range of different manipulation tasks.

\subsection{Reinforcement Learning in Robotics}

Reinforcement Learning (RL) \cite{sutton1998reinforcement,kormushev2013reinforcement} has been applied in robotics, as it promises a way to learn complex actions on complex robotic systems by just providing informing the robot whether its actions were successful (positive reward) or not (negative reward).
\cite{peters2003reinforcement} reviewed some of the RL concepts in terms of applicability to control complex humanoid robots and highlighting some of the issues with greedy policy search and gradient based methods.
How to generate the right reward is an active topic of research. Intrinsic motivation and curiosity have been shown to provide means to explore large state spaces, such as the ones found on complex humanoids, faster and more efficient \cite{frank2014curiosity}.

\subsection{Deep Visuomotor Policies}

To enable robots to learn manipulation skills with little prior knowledge, a convolutional neural network (CNN) based policy representation architecture (deep visuomotor policies) and its guided policy search method were introduced by Sergey et al. \cite{levine2015end,levine2015learning}. The deep visuomotor policies map joint angles and camera images directly to the joint torques. Robot configurations are the only necessary prior knowledge. The policy search method consists of two phases, i.e., optimal control phase and supervised learning phase. The training consists of three procedures, i.e., pose CNN training, trajectories pre-training, and end-to-end training.

The deep visuomotor policies did enable robots to learn manipulation skills with little prior knowledge through supervised learning, but pre-collected datasets were necessary. Human involvements in the datasets collection made this method less autonomous. Besides, the training method specifically designed to speed up the contact-rich manipulation learning made it less flexible for other manipulation tasks.

\begin{figure*}[tb!]
  \centering
  \includegraphics[width=2.05\columnwidth]{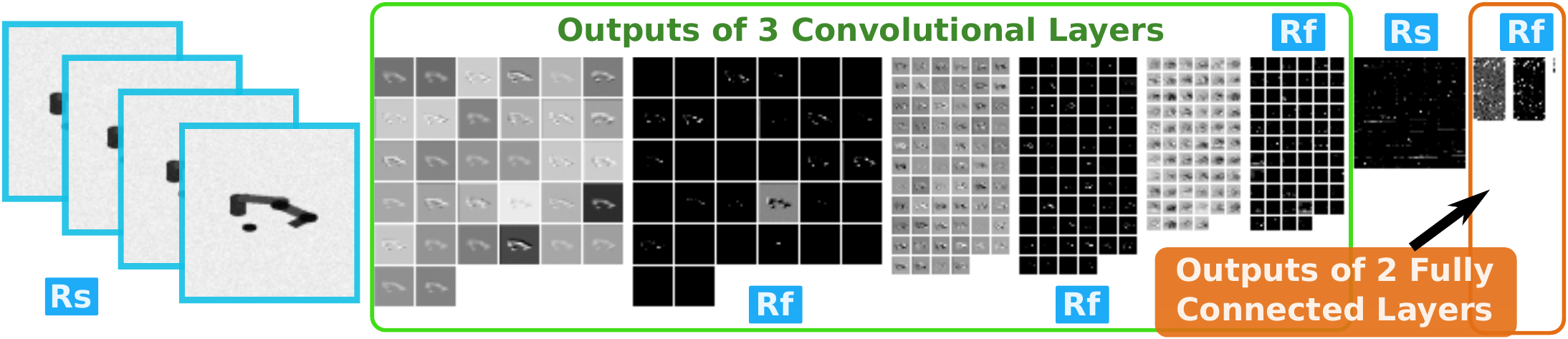}
\caption{Schematic of the DQN layers for end-to-end learning and their respective outputs.
Four input images are reshaped (Rs) and then fed into the DQN network as grey-scale images (converted from RGB).
The DQN, consists of three convolutional layers with rectifier layers (Rf) after each, followed by a reshaping layer (Rs) and two fully connected layers (again with a rectifier layer in between).
The normalized outputs of each layer are visualized. (Note: The outputs of the last four layers are shown as matrices instead of vectors.)}
\label{fig:nn_output}
\end{figure*}

\subsection{Deep Q Network}
The DQN, a topical example of DRL, satisfies both the autonomy and flexibility requirements for learning from exploration. It successfully learnt to play 49 different Atari 2600 games, achieving a human-level of control \cite{mnih2015human}. The DQN used a deep convolutional neural network (CNN) \cite{krizhevsky2012imagenet} to approximate  a Q-value function. It maps raw pixel images directly to actions. No pre-input feature extraction is needed. 
The only one thing is to let the algorithm improve policies through playing games over and over again. It learnt playing 49 different games, using the same network architecture with no modification. 

The DQN is defined by its inputs -- raw pixels of game video frames and received rewards -- and outputs, i.e., the number of available actions in a game \cite{mnih2015human}. This number of actions is the only prior knowledge, which means no robot configuration information is needed to the agent, when using the DQN for motion control. However, in the DQN training process, the Atari 2600 game engine worked as a reward function, but for robotic motion control, no such engine exists. To apply it in robotic motion control, a reward function is needed to assess trials. Besides, sensing noise and higher complexity and dynamics are inevitable issues for real-world applications.

\section{Problem Definition and System Description}

A common problem in robotic manipulation is to reach for the object to be interacted with. This target reaching task is defined as controlling a robot arm, such that its end-effector is reaching a specific target configuration. We are interested in the case in which a robot performs the target reaching with visual perception only. To learn such a task, we developed a system consisting of three parts:
\begin{itemize}
\item a 2D simulator for robotic target reaching, creating the visual inputs to the learner
\item a deep reinforcement learning framework based on the DQN implementation by Google Deepmind \cite{mnih2015human}, and
\item a component of ROS-based interfaces to control a Baxter robot according to the DQN outputs.
\end{itemize}

\subsection{DQN-based Learning System} 
\label{sec:system}
\begin{figure}[tb!]
  \centering
  \includegraphics[width=0.95\columnwidth]{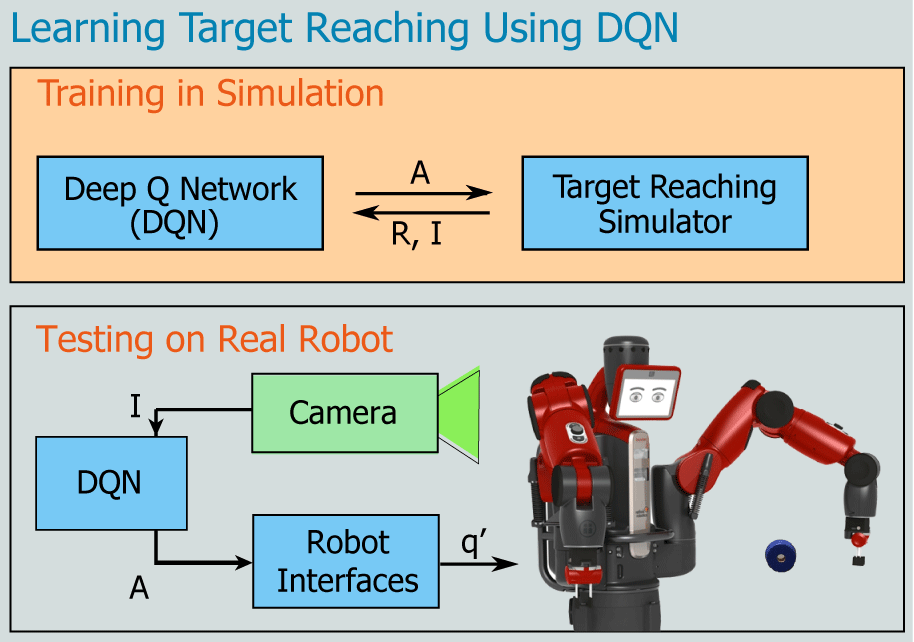}
\caption{System overview} \label{fig:framework}
\end{figure}

The DQN adopted here has the same architecture with that for playing Atari games, which contains three convolutional layers and two fully connected layers \cite{mnih2015human}. Its implementation is based on the Google Deepmind DQN code\footnote{https://sites.google.com/a/deepmind.com/dqn/} with minor modifications. Fig.~\ref{fig:nn_output} shows the architecture and examplary output of each layer. The inputs of the DQN include rewards and images. Its output is the index of the action to take.
The DQN learns target reaching skills in the interactions with the target reaching simulator. 
An overview of the system framework for both the learning in simulation and testing on a real robot is shown in Fig.~\ref{fig:framework}.

When training or testing in simulation, the target reaching simulator provides the reward value ($R$) and image ($I$). $R$ is used for training the network. The action output ($A$) of the DQN is directly sent to the simulated robotic arm. 

When testing on a Baxter robot using camera images, an external camera provides the input images ($I$). The action output ($A$) of the DQN is implemented on the robot controlled by ROS-based interfaces. The interfaces control the robot by sending updated robot's poses ($q'$).

\subsection{Target Reaching Simulator}
\label{sec:task_def}

\begin{figure}[t!]
\begin{center}
\subfigure[Schematic diagram]
{
\includegraphics[width=0.4\columnwidth]{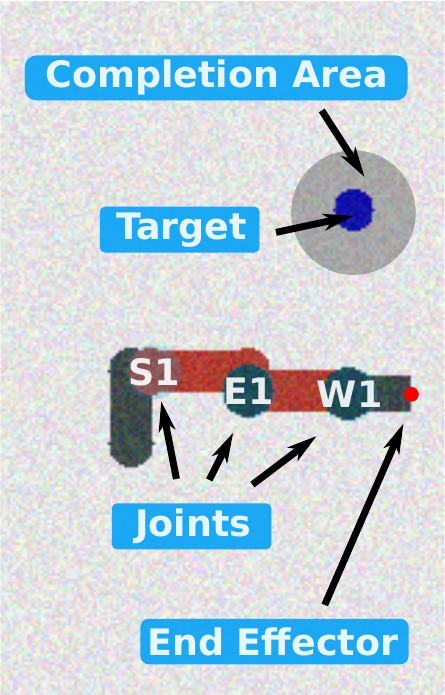}
\label{fig:task_schematic}
}
\subfigure[The robot simulator during a successful reach]
{
\includegraphics[width=0.4\columnwidth]{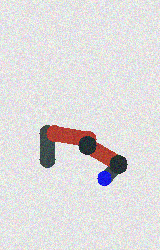}
\label{fig:simulator}
}
\end{center}
\vspace{-0.4cm}
\caption{The 2D target reaching simulator, providing visual inputs to the DQN learner. It was implemented from scratch, no simulation platform was used.}
\label{fig:reaching_task}
\end{figure}

We simulate the reaching task to control a three-joint robotic arm in 2D (Fig. \ref{fig:reaching_task}). The simulator was implemented from scratch. In the implementation, no simulation platform was used. As shown in Fig. \ref{fig:task_schematic}, the robotic arm consists of four links and three joints, whose configurations are consistent to the specifications of a Baxter arm, including joints constraints. The blue spot is the target to be reached. For a better visualization, the position of the end-effector is marked with a red spot. The simulator can be controlled by sending specific commands to the individual joints ``S1'', ``E1'' and ``W1''. The simulator screen resolution is $160\times320$.

The corresponding real scenario that the simulator simulates is: with appropriate constant joint angles of other joints on a Baxter arm, the arm moves in a vertical plane controlled by joints ``S1'', ``E1'' and ``W1'', and a controller (game player) observes the arm through an external camera placed directly aside it with a horizontal point of view. The three joints are in position control mode. The background is white.

In the system, the 2D simulator is used as a target reaching video game in connection with the DQN setup. It provides raw pixel inputs to the network and has nine options for action, i.e., three buttons for each joint: joint angle increasing, decreasing and hold. The joint angle increasing/decreasing step is constant at 0.02 rad. At the beginning of each round, joints ``S1'', ``E1'' and ``W1'' will be set to a certain initial pose, such as [0.0, 0.0, 0.0] rad; and the target will be randomly selected. 

In the game playing, a reward value will be returned for each button press. The reward value is determined by a reward function introduced in Section \ref{sec:reward}. When satisfying some conditions, the game will terminate. The game terminal is determined by the reward function as well. For a player, the goal is to get an as high as possible accumulated reward before the game terminates. For clarity, we name 
an entire trial from the start of the game to its terminal as one round.

\subsection{Reward Function}
\label{sec:reward}

\begin{algorithm}[tpb!]
 \SetKwFunction{ComputeDistance}{ComputeDistance}
 \SetKwInOut{Input}{input}\SetKwInOut{Output}{output}
 \Input{$P_t$: the target 2D coordinates\;\\
  $P_e$: the end-effector 2D coordinates.}
 \Output{$R$: the reward for current state\;\\
 $T$: whether the game is terminal.}
 \BlankLine

$Dis$ = \ComputeDistance{$P_t,\ P_e$}\;
$DisChange=Dis-PreviousDis$\;
\uIf{$DisChange > 0$}{
	$R=-1$;
}
\uElseIf{$DisChange < 0$}{
	$R=1$;
}
\Else{
	$R=0$;
}
$R_{acc}= R_t+R_{t-1}+R_{t-2}$\;
\uIf{$R_{acc} < -1$}{
	$T=True$;
}
\Else{
	$T=False$;
}

\caption{Reward Function}
\label{alg:reward_function}
\end{algorithm}

To keep consistent to the DQN setup, the reward function has two return values: one for the reward of each action; the other shows whether the target reaching game is terminal. Its algorithm is shown in Algorithm \ref{alg:reward_function}. The reward of each action is determined according to the distance change between the end-effector and the target. If the distance gets closer, the reward function returns 1; if gets further, returns -1; otherwise returns 0. If the sum of the latest three rewards is smaller than -1, the game terminates. This reward function was designed as a first step, more study is necessary to get an optimal reward function.

\section{Experiments and Results}
To evaluate the feasibility of the DQN-based system in learning performing target reaching, we did some experiments in both simulation and real-world scenarios. The experiments consist of three phases: training in simulation, testing in simulation, and testing in the real world.

\subsection{Training in Simulation Scenarios}
\label{sec:training}

\begin{figure*}[tb!]
\begin{center}
\subfigure[Settings A: simulation images]
{
\includegraphics[width=0.37\columnwidth]{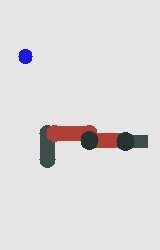}
\label{fig:setting_a}
}
\subfigure[Setting B: simulation images + noise]
{
\includegraphics[width=0.37\columnwidth]{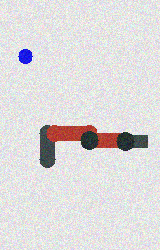}
\label{fig:setting_b}
}
\subfigure[Setting C: Setting B + random initial pose]
{
\includegraphics[width=0.37\columnwidth]{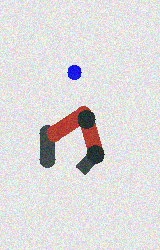}
\label{fig:setting_c}
}
\subfigure[Setting D: Setting C + random image offset]
{
\includegraphics[width=0.37\columnwidth]{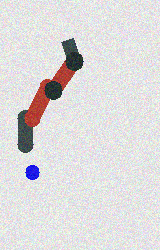}
\label{fig:setting_d}
}
\subfigure[Setting E: Setting D + random link length]
{
\includegraphics[width=0.37\columnwidth]{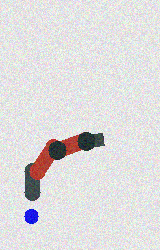}
\label{fig:setting_e}
}
\end{center}
\vspace{-0.32cm}
\caption{Screenshots highlighting the different training scenarios for the agents.} 
\label{fig:sim_setting_screenshot}
\end{figure*}

To evaluate the capability of the DQN to adapt to some noise commonly concerned in robotic manipulation, 
we trained several agents with different simulator settings. The different settings include sensing noise, image offsets, variations in initial arm pose and link length. The setting details for training the five agents are shown in Table \ref{tab:settings}. Their screenshots are shown in Fig. \ref{fig:sim_setting_screenshot}, respectively.

\begin{table}[htpb!]
\caption{Agents and training settings}
\label{tab:settings}
\centering	
\renewcommand\arraystretch{1.1}
\begin{tabular}{c | l}	
\toprule 
\toprule

\bfseries Agent & \bfseries Simulator Settings \\ 

\midrule 
A & constant initial pose\\
B & Setting A + random image noise\\
C & Setting B + random initial pose\\
D & Setting C + random image offset\\
E & Setting D + random link length\\

\bottomrule
\bottomrule 
\end{tabular}
\end{table}

Agent A was trained in Setting A where the 2D robotic arm was initialized to the same pose ([0.0, 0.0, 0.0] rad) at the beginning of each round. There was no image noise in Setting A. To simulate camera sensing noise, random noise was added in Setting B, on the basis of Setting A. The random noise was with a uniform distribution with a scale between -0.1 and 0.1 (for float pixel values).

In Setting C, in addition to random image noise, the initial arm pose was randomly selected. In the training of Agent D, random image offsets were added on the basis of Setting C. The offset ranges in u and v directions were respectively [-23, 7] and [-40, 20] in pixel. Agent E was trained with dynamic arm link lengths. The link length variation ratio was [-4.2, 12.5]\% with respect to the link length settings in the previous four settings. The image offsets and link lengths were randomly selected at the beginning of each round, and stayed unchanged in the entire round (not vary at each frame). All the parameters for noisy factors were empirically selected as a first step.

All the agents were trained using more than 4 million steps within 160 hours. Due to the difference in setting complexity, the time-cost for the simulator to update each game video frame varies in five different settings. Therefore, within 160 hours, the exact numbers of used training steps for the five agents are different. They are 6.475, 6.275, 5.225, 4.75 and 6.35 million, respectively.

\begin{figure}[tb!]
  \centering
  \includegraphics[width=0.9\columnwidth]{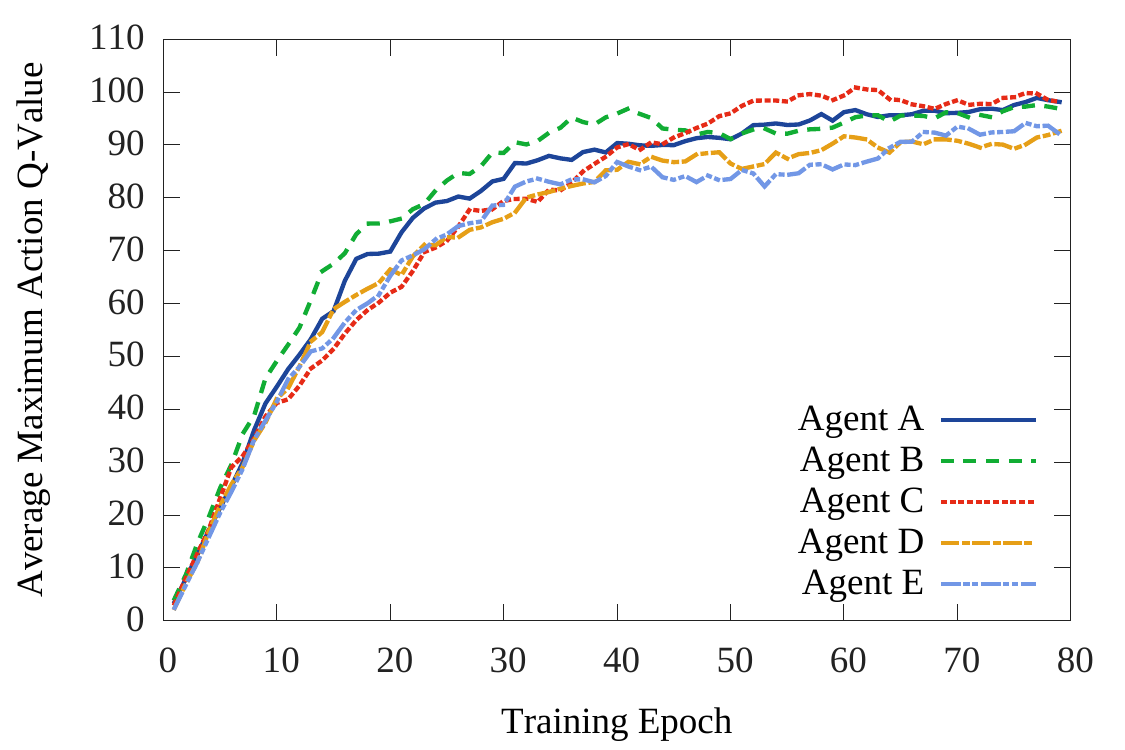}
  \vspace{-0.15cm}
\caption{Action Q-value converging curves. Each epoch contains 50,000 training steps. The average maximum action Q-values are the average of the estimated maximum Q-values for all states in a validation set. The validation set has 500 frames.} \label{fig:q_value}
\end{figure}

The action Q-value converging curves are shown in Fig. \ref{fig:q_value}. The Q-value curves are respect to training epochs. Each epoch contains 50,000 training steps. Fig. \ref{fig:q_value} shows the converging case before 80 epochs, i.e., 4 million training steps. The average maximum action Q-values are the average of the estimated maximum Q-values for all states in a validation set. 
The validation set was randomly selected at the beginning of each training.

From Fig. \ref{fig:q_value}, we can observe that all the five agents converge towards to a certain Q-value state, although their values are different. One thing we have to emphasize is this converging is just for average maximum action Q-values. A high value might but not necessarily indicate a high performance of an agent in performing target reaching, since this value cannot completely indicate the target reaching performance.

\subsection{Testing in Simulation Scenarios}

We tested the five agents in simulation scenarios with the 2D simulator. Each agent was tested in all those five settings in Table \ref{tab:settings}. Each test took 200 rounds, i.e., terminated 200 times. More testing rounds can make the testing results closer to the ground truth, but need too much time.

In the testing, task success rates were evaluated. In the computation of success rates, it is regarded as a success when the end-effector gets into a completion area with a radius of 16 cm around a target, as shown in the grey circle in Fig. \ref{fig:task_schematic}, which is twice size of the target circle. The radius of 16 cm is equivalent to 15 pixels in the simulator screen. However, for the DQN, this completion area is a ellipse (a=8 pixels, b=4 pixels), since the simulator screen will be resized from $160\times320$ to $84\times84$ before being input to the learning core.

Table \ref{tab:success_rate1} shows the success rates of different agents in different settings after 3 million training steps (60 epochs). The data in the diagonal (with a cell color of gray) shows the success rate of each agent tested with the same setting in which it trained, i.e., Agent A was tested in Setting A. 

We also did some experiments for agents from different training steps. Table \ref{tab:success_rate2} shows the success rates of different agents after some certain training steps. The success rates of each agent were tested with the same simulator setting in which it was trained, i.e., the case in the diagonal of Table \ref{tab:success_rate1}. In Table \ref{tab:success_rate2}, ``f'' indicates the final number of steps used for training each agent in 160 hours, as mentioned in Section \ref{sec:training}.

What we will discuss regarding the data in Table \ref{tab:success_rate1} and \ref{tab:success_rate2} is based on the assumption that some outliers of some conclusions appeared accidentally due to the limited number of testing rounds. Although 200 testing rounds are already able to extract data changing trends in success rates, they are insufficient to extract the ground truth. Some minor success rate distortions happen occasionally. To make the conclusions more convincing, more study is necessary.

From Table \ref{tab:success_rate1}, we can find that Agent A and B can both adapt to Setting A and B, but can not adapt to the other three settings. This shows that these two agents are robust to random image noise, but not robust to dynamic arm initial pose and link length, and image offsets. The random image noise is not a key feature in these two agents.

In addition, other than the settings in which they are trained, Agent C, D and E can also achieve relatively high success rates in the settings with fewer noisy factors than their training settings. This indicates that agents trained with more noisy factors can adapt to settings with fewer noisy factors.

\begin{table}[tb!]
\caption{Success rates (\%) in different settings}
\label{tab:success_rate1}
\centering	
\begin{tabular}{c |c | c | c | c | c}	
\toprule
\toprule

\multirow{2}{*}{Agent} & 

\multicolumn{5}{c}{Setting} 
\\ 

\ & \bfseries A & \bfseries B & \bfseries C & \bfseries D
& \bfseries E
\\ 

\midrule
\bfseries A & \cellcolor{gray}51.0 & 53.0 	& 14.0 & 8.5 & 8.5 
\\
\bfseries B & 50.5 & \cellcolor{gray}49.5 & 11.0 & 8.0 & 10.0
\\
\bfseries C & 32.0 & 34.5 & \cellcolor{gray}36.0 & 22.5 & 14.0
\\
\bfseries D & 13.5 & 16.5 & 22.0 & \cellcolor{gray}19.5 & 15.0 
\\
\bfseries E & 13.0 & 16.5 & 20.0 & 16.5 & \cellcolor{gray}19.0 
\\
\bottomrule 
\bottomrule 
\end{tabular}
\end{table}

\begin{table}[tb!]
\caption{Success rates (\%) after different training steps}
\label{tab:success_rate2}
\centering	
\begin{tabular}{c |c | c | c | c | c}	
\toprule
\toprule

\multirow{2}{*}{Agent/Setting} & 

\multicolumn{5}{c}{Training Steps / million} 
\\ 

\ & \bfseries 1 & \bfseries 2 & \bfseries 3 & \bfseries 4
& \bfseries f
\\ 

\midrule
\bfseries A & 36.0 & 43.0 & 51.0 & 36.0 & 36.5 
\\
\bfseries B & 58.0 & 55.5 & 49.5 & 51.5 & 13.5
\\
\bfseries C & 30.5 & 33.0 & 36.0 & 48.0 & 13.5
\\
\bfseries D & 16.5 & 17.5 & 19.5 & 26.5 & 14.0 
\\
\bfseries E & 13.0 & 18.5 & 19.0 & 23.0 & 27.0 
\\
\bottomrule 
\bottomrule 
\end{tabular}
\end{table}

In Table \ref{tab:success_rate2}, we can find that the success rate of each agent normally goes up after more training steps. This shows that, in the training process, all the five agents can learn to adapt to the noisy factors presented in their settings. However, some goes down after a certain training step, e.g., the success rate of Agent A goes down after 4 million training steps. Theoretically, with a appropriate reward function, the DQN should perform better and better, and the success rates should go up iteratively. 

The going down case was quite possibly caused by the reward function, which has the possibility to guide the agent to a wrong direction. For the case in this paper, the evaluation is based on success rates, but the reward function is based on distance changes. The relation between success rates and distance changes is indirect. This indirect relation provides the incorrect guidance possibility. This should be considered carefully in future work.

Table \ref{tab:success_rate2} also shows that the success rate of the agent trained in a more complicated setting is normally smaller than that in a simpler setting, and needs more training time to get to a same level of success rate. For example, the success rate of Agent E is smaller than that of Agent D in each training episode, but is close to that of Agent D in a latter training episode.

In general, no matter whether the discussion assumption holds or not, the data in Table \ref{tab:success_rate1} and \ref{tab:success_rate2} at least shows that the DQN has the capability to adapt to these noisy factors, and is feasible to learn performing target reaching from exploration in simulation. However, more study is necessary to increase the success rates.

\subsection{Real World Experiment Using Camera Images}
\label{sec:real_validation}

To check the feasibility of trained agents in the real world, we did a target reaching experiment in real scenarios using camera images, i.e., the second phase mentioned in Section \ref{sec:system}. In this experiment, we used Agent B trained with 3 million steps, which has relatively high success rates for both Setting A and B in the testing in simulation.

\begin{figure}[tb!]
\begin{center}
\subfigure[Real world experiment using camera images]
{
\includegraphics[width=0.52\columnwidth]{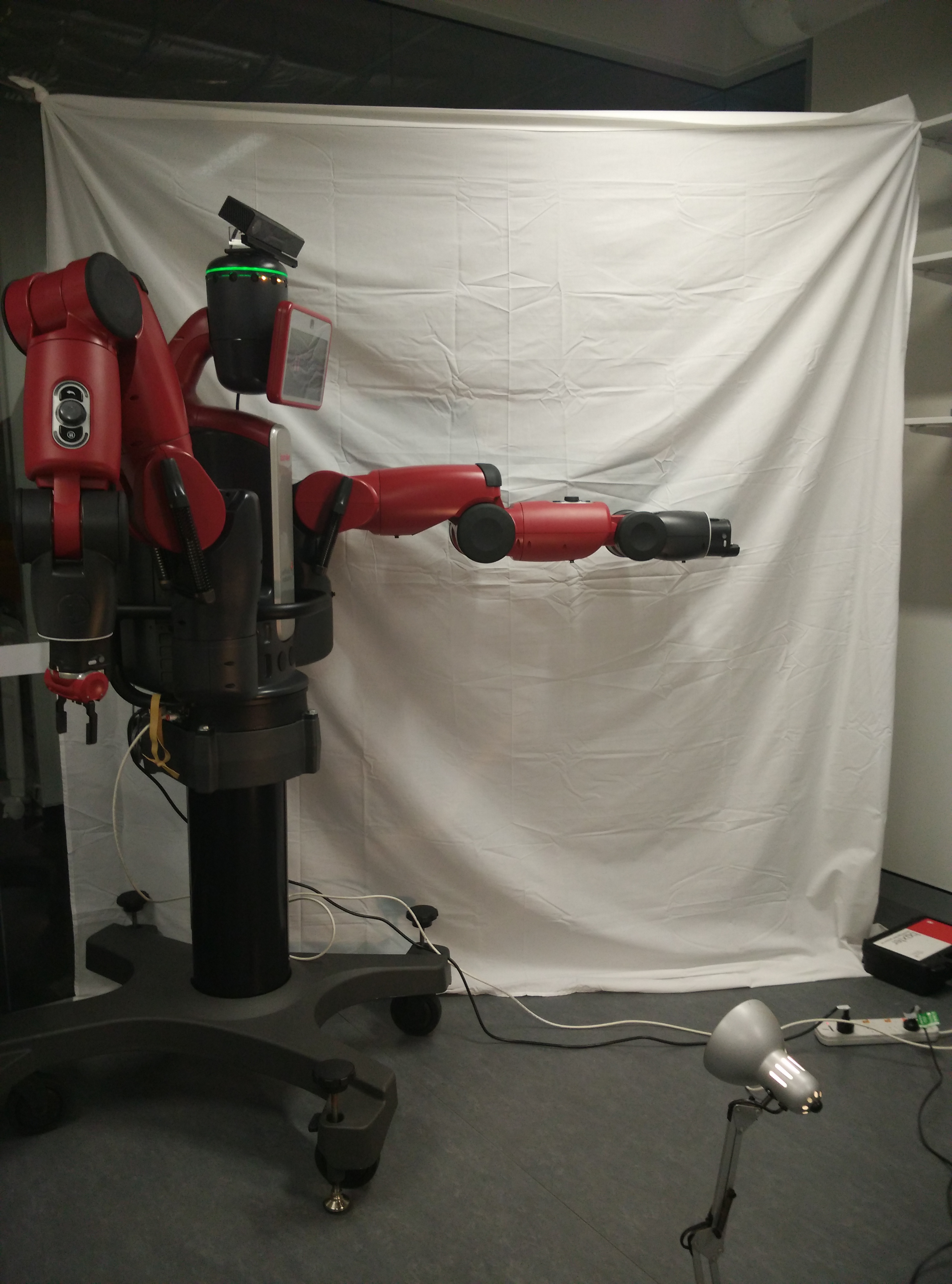}
\label{fig:test_scene1}
}
\subfigure[A sample input image]
{
\includegraphics[width=0.35\columnwidth]{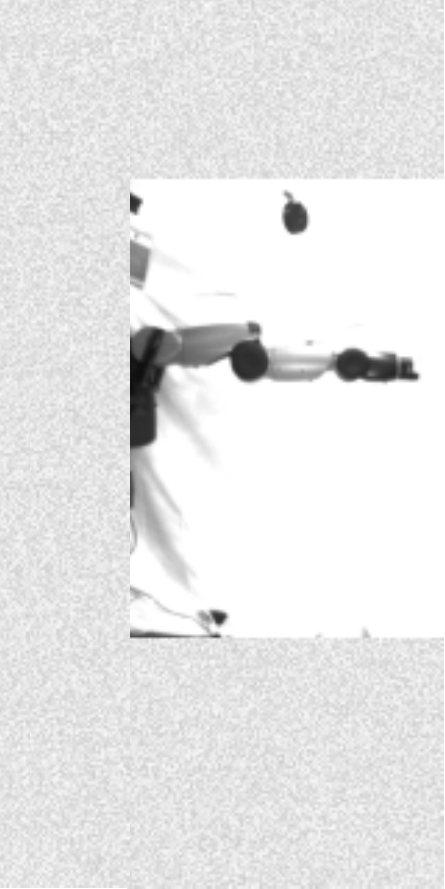}
\label{fig:crop}
}
\end{center}
\vspace{-0.3cm}
\caption{Testing scene and a sample input of the real world experiment using camera images. In the testing scene, a Baxter arm moved on a vertical plane with a white background. To guarantee that images input to the DQN have an as consistent as possible appearance to those in simulation scenarios, camera images were cropped and masked with a boundary. The boundary is from the background of a simulator screenshot.}
\label{fig:validation1}
\end{figure}

The experiment settings were arranged to the case that the 2D simulator simulated, i.e., a Baxter arm moved on a vertical plane with a white background. A grey-scale camera was placed in front of the arm, observing the arm with a horizontal view of point (for the DQN, the grey-scale camera is the same with a color camera, since even the images from Atari games and the 2D target reaching simulator are RGB-color images, they are converted to grey-scale images prior to being input to the network). The background was a white sheet. The testing scene and a sample input to the DQN are shown in Fig. \ref{fig:test_scene1} and \ref{fig:crop}, respectively.

In the experiment, to make the agent work in the real world, we tried to match the arm position (in images) in real scenarios to that in simulation scenarios. The position adjustment was made through changing camera pose and image cropping parameters. However, no matter how we adjusted, it did not reach the target. The success rate is 0.

Other than the success rate, we also got a qualitative result: Agent B mapped specific input images to certain actions, but the mapping was ineffective for performing target reaching. There were some kind of mapping distortions between real and simulation scenarios. The distortions might be caused by the differences between real-scenario and simulation-scenario images.

\subsection{Real World Experiment Using Synthetic Images}
\label{sec:fake_validation}
To verify the analysis regarding the reason why Agent B failed to perform target reaching, 
we did another real world experiment using synthetic images instead of camera images. In the experiment, the synthetic images were generated by the 2D simulator according to real-time joint angles (``S1'', ``E1'' and ``W1'') on a Baxter robot. The real-time joint angles were provided by the ROS-based interfaces. In this case, there was no difference between real-scenario and simulation-scenario images. 
All other settings were the same with those in Section \ref{sec:real_validation}, as shown in Fig. \ref{fig:baxter}.

In this experiment, we used the same agent that was used in Section \ref{sec:real_validation}, i.e., Agent B trained with 3 million steps. It achieved a consistent success rate with that in the simulation-scenario testing. 

According to the results, we can conclude that the reason why Agent B failed in completing the target reaching task with camera images is the existence of input image differences. These differences might come from camera pose variations, color and shape distortions, or some other factors. More study is necessary to exactly figure out where the differences came from.

\section{Conclusion and Discussion}
The DQN-based system is feasible to learn performing target reaching from exploration in simulation, using only visual observation with no prior knowledge. 
However, the agent (Agent B) trained in simulation scenarios failed to perform target reaching in the real world experiment using camera images as inputs. Instead, in the real world experiment using synthetic images as inputs, the agent got a consistent success rate with that in simulation. These two different results show that the failure in the real world experiment with camera images was caused by the input image differences between real and simulation scenarios. To determine the causes of these more work is required.

In the future, we are looking at either decreasing the image differences or making agents robust to these differences. 
Decreasing the differences is a trade-off between making the simulator more consistent to real scenarios and preprocessing input images to make them more consistent to those in simulation scenarios. If choose to increase the fidelity of the simulator, it will most likely result in a slow-down of the simulation, increasing training time.

Regarding making agents robust to the differences, there are four possible methods: adding variations of the factors causing the image differences into simulation scenarios when training, adding a fine-tuning process in real scenarios after the training in simulation scenarios, training in real scenarios directly, and designing a new DRL architecture (still can be a DQN) which is robust to the image differences.

In addition to solving the problem of image differences, more study is necessary in the design of reward function. 
A good reward function is the key to get effective motion control or even manipulation skills and also speed up the learning process. The reward function used in this work is just a first step. It is far less than enough to be a good reward function. Other than the effectiveness and efficiency concerns, a good reward function needs also to be flexible to a range of general purpose motion control or even manipulation tasks.

Besides, the visual perception in this work is from an external monocular camera. An on-robot stereo camera or RGBD sensor can be a more effective and practical solution for applications in the 3D real world. The joint control mode in this work is position control, some other control modes like speed control and torque control are more common and appropriate for dynamic motion control and manipulation in real-world applications.

\section*{Acknowledgements}
This research was conducted by the Australian Research Council Centre of Excellence for Robotic Vision (project number CE140100016). Computational resources and services used in this work were partially provided by the HPC and Research Support Group, Queensland University of Technology (QUT).

\bibliographystyle{named}
\bibliography{deep_manipulation,rl}

\end{document}